%% file: main.tex
\documentclass[sigplan]{acmart}
\AtBeginDocument{%
  }

\setcopyright{acmlicensed}
\copyrightyear{2018}
\acmYear{2018}
\acmDOI{XXXXXXX.XXXXXXX}

\acmConference[Conference acronym 'XX]{Make sure to enter the correct
  conference title from your rights confirmation emai}{June 03--05,
  2018}{Woodstock, NY}
\acmISBN{978-1-4503-XXXX-X/18/06}

\usepackage{fancyhdr}

\newcommand{\equalcontrib}{\textsuperscript{\textasteriskcentered}}
\newcommand{\correspondingauthor}{\textsuperscript{\textdagger}}

\begin{document}

\title{\texttt{GCoder}: Improving Large Language Model for Generalized Graph Problem Solving}


\author{Qifan Zhang\equalcontrib\textsuperscript{1}, Xiaobin Hong\equalcontrib\textsuperscript{1,2}, Jianheng Tang\equalcontrib\textsuperscript{1}, Nuo Chen\textsuperscript{1} \\
Yuhan Li\textsuperscript{1}, Wenzhong Li\textsuperscript{2}, Jing Tang\textsuperscript{1}, Jia Li\textsuperscript{1}\correspondingauthor}
\affiliation{%
  \institution{\textsuperscript{1}Hong Kong University of Science and Technology (GuangZhou)}
  \city{Guangzhou}
  \country{China}
}
\affiliation{%
  \institution{\textsuperscript{2}State Key Laboratory for Novel Software Technology, Nanjing University}
  \city{Nanjing}
  \country{China}
}

\email{qzhang297@connect.hkust-gz.edu.cn, xiaobinhong@smail.nju.edu.cn}
\email{jtangbf@connect.ust.hk, jialee@ust.hk}

\thanks{\equalcontrib Equal contribution.}
\thanks{\correspondingauthor Corresponding author.}

\renewcommand{\shortauthors}{Trovato et al.}

\newcommand{\ourmodel}{\texttt{GCoder} }
\newcommand{\ourdataset}{\texttt{GraphWild} }
\begin{abstract}
Large Language Models (LLMs) have demonstrated strong reasoning abilities, making them suitable for complex tasks such as graph computation. Traditional reasoning steps paradigm for graph problems is hindered by unverifiable steps, limited long-term reasoning, and poor generalization to graph variations. To overcome these limitations, we introduce \texttt{GCoder}, a code-based LLM designed to enhance problem-solving in generalized graph computation problems. Our method involves constructing a extensive training dataset, \texttt{GraphWild}, featuring diverse graph formats and algorithms. We employ a multi-stage training process, including Supervised Fine-Tuning (SFT) and Reinforcement Learning from Compiler Feedback (RLCF), to refine model capabilities. For unseen tasks, a hybrid retrieval technique is used to augment performance. Experiments demonstrate that \texttt{GCoder} outperforms GPT-4o, with an average accuracy improvement of \textbf{16.42\%} across various graph computational problems. Furthermore, \ourmodel efficiently manages large-scale graphs with millions of nodes and diverse input formats, overcoming the limitations of previous models focused on the reasoning steps paradigm. This advancement paves the way for more intuitive and effective graph problem-solving using LLMs. Code and data are available at here\footnote{ https://github.com/Bklight999/WWW25-GCoder/tree/master.}.
\end{abstract}



\maketitle

\input{intro1}

\section{Methodology}
\begin{figure}[t]
    \centering
    \includegraphics[width=\linewidth]{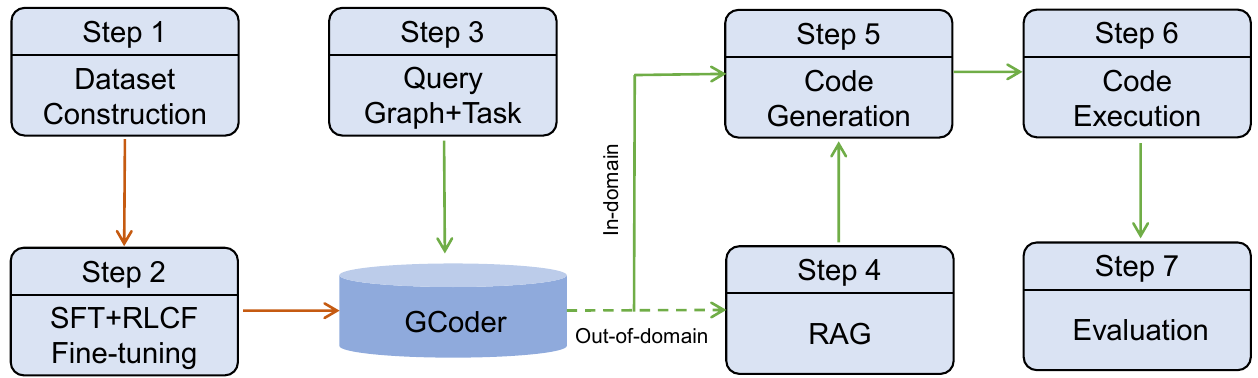}
    \vspace{-3mm}
    \caption{The workflow of our proposed \texttt{GCoder}.}
    \vspace{-6mm}
    \label{fig:pipeline}
\end{figure}

Figure \ref{fig:pipeline} provides an overview of the \ourmodel workflow, which can be divided into three parts: \ourdataset Dataset Construction (Step 1), Multi-stage Fine-tuning (Step 2), and Model Inference (Steps 3-7).
In the Dataset Construction phase, we create the training dataset by combining our graph and algorithm samples with key-joint among two databases, which outputs our \ourdataset dataset composed of \texttt{Problem-Code} pairs with diverse graph input formats and tasks.
During the Multi-stage Fine-tuning, we perform supervised fine-tuning to train our model for code generation. Subsequently, we employ the RLCF technique for preferred code alignment.
In the Model Inference, we check whether the query task seen in model fine-tuning,  determines 
\texttt{In-domain} tasks or \texttt{Out-of-domain} tasks. For in-domain tasks, we directly prompt \ourmodel to generate, execute, and evaluate the code. For out-of-domain tasks, we use RAG to retrieve relevant code documents before inference, enhancing the generalization of \texttt{GCoder} for unseen tasks.

\subsection{Training Dataset Construction}\label{Dataset Construction}
Unlike previous work, which includes datasets with only fixed-format graph inputs and a limited number of tasks, we introduce \ourdataset, which features dozens of graph input formats and hundreds of graph tasks. 

\begin{figure*}[t]
\centering
\includegraphics[width=\textwidth]{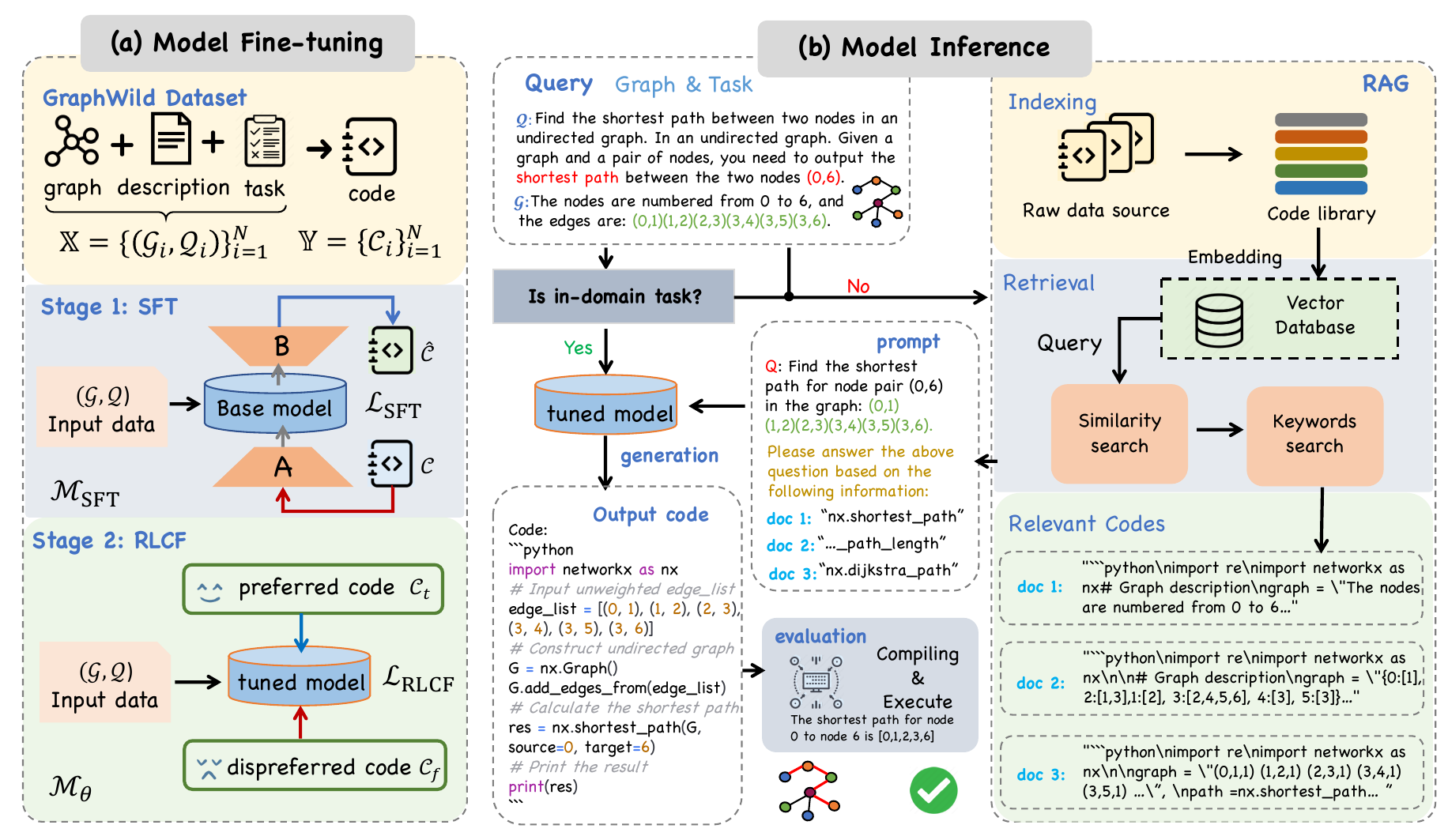}
\caption{The overview framework of \texttt{GCoder}, which consists of (a) Model Fine-tuning and (b) Model Inference two pipelines. In model fine-tuning, we develop SFT and RLCF two fine-tuning stages with our constructed GraphWild dataset. In model inference, the in-domain query task is directly prompt our tuned model for code generation, while the out-of-domain task is enhanced by the RAG technique. We execute the generated code and evaluate the code output with a ground-truth answer.}
\label{RAG_inference}
\end{figure*}



\noindent\textbf{Graphs Dataset.} To create a dataset with a diverse graph input format, we construct our graph dataset, which includes graphs with natural language descriptions of two types: graphs of traditional data structure and graphs of real-life scenarios.

\textbf{$\bullet$ Graphs of Traditional Data Structure $\mathcal{G}_d$:} We construct graphs using traditional data structures, such as adjacency matrix, adjacency list, and edge list, through the following method: First, we employ the Erdős-Rényi (ER) model ~\cite{erdHos1960evolution} to generate random graphs. This model requires two parameters: the number of nodes $n$ and the probability $p$ of an edge existing between any two nodes. For each node pair, an edge is randomly created with probability $p$. Next, we generate natural language descriptions for random graphs based on different templates corresponding to their topology structures. Note that to accommodate a wider range of graph algorithms, we generate different kinds of graphs, including \textit{undirected graphs, directed graphs, bipartite graphs}, and \textit{weighted graphs}. 

    
\textbf{$\bullet$ Graphs of Real-Life Scenarios $\mathcal{G}_r$:} We create graphs with real-life scenarios description. We first generate random ER graphs, we then prompt two close-sourced LLMs(i.e., GPT-4o and Deepseek) to generate diverse real-world graphs based on these ER graphs across various domains, including \textit{Computer Science, Data, Bioinformatics, Finance, Logistics and Transportation, Chemistry, Web analysis, and Physics}.

After the above processing, we obtain a set of graphs with diverse input formats, denoted as $\mathcal{G} = (\mathcal{G}_d, \mathcal{G}_r)$. 
We then utilize GPT-4o to generate various methods for extracting graphs from natural language descriptions. For $\mathcal{G}_d$, we design regular expressions tailored to each graph for extraction using Python code. In the case of $\mathcal{G}_r$, we develop step-by-step reasoning analyses to obtain the graphs.
The final forms of $\mathcal{G}_d$ and $\mathcal{G}_r$ are defined as $\mathcal{G}_d = \mathcal{(N_G,C_G,T_G)}$ and $\mathcal{G}_r = \mathcal{(N_G,R_G,T_G)}$, respectively. Here, $\mathcal{N_G}$ denotes the natural language description, $C_{\mathcal{G}}$ is the code used to read the graph, $\mathcal{R_G}$ is the reasoning process applied to interpret the graph, and $\mathcal{T_G}$ specifies the type of the graph (e.g., an undirected graph).

\noindent\textbf{Algorithms Dataset.} Our objective is to develop a model proficient in handling a wide array of graph computational problems. To achieve this, we have compiled a comprehensive dataset encompassing numerous graph algorithms. We sourced documents from NetworkX~\cite{hagberg2008exploring}, a leading library for graph algorithms. This effort resulted in 313 documents covering a variety of graph-related problems.
Additionally, we utilized GPT-4o to determine the type of graph associated with each document, ensuring proper alignment with its corresponding graph type. The final structure of the algorithm dataset is defined as $\mathcal{A} = (\mathit{D},\mathcal{T_G})$, where $\mathit{D}$ is a NetworkX document containing the algorithm's input parameters, a description of the problem and algorithm, and a code example. $\mathcal{T_G}$ represents the type of graph relevant to the document.


\noindent\textbf{Training Dataset.} Our final training dataset is a combination of the graphs dataset and the algorithms dataset. Specifically, these datasets can be viewed as two tables that we join based on their common attribute $\mathcal{T_G}$. After connection, we obtain a table containing the following two types of data:
\begin{equation}
\begin{split}
    \mathcal{D}_d = \{\mathcal{(N_G,C_G,T_G}, D)\}, \quad
    \mathcal{D}_r = \{\mathcal{(N_G,R_G,T_G}, D)\},
\end{split}
\end{equation}
We then prompt GPT-4o to generate graph problems and their corresponding code, forming the final format of the training data.





After processing, we initially obtained a dataset of 68,597 entries. However, not all data points are usable due to issues such as incorrect outputs or execution failures. To ensure the effectiveness of our training data, we implement a thorough data-cleaning process. We first check each code can be compiled and executed without any errors.
Subsequently, both GPT-4o and human experts offer review for programming outputs to verify their correctness. Only those codes that are executable and produce correct results will be retained, resulting in a dataset of \textbf{45,236} data points.

\noindent\textbf{Data Augmentation.} Although our initial training dataset, generated by LLMs, encompasses over 300 algorithms for graph problems, it still faces several issues.

\textbf{$\bullet$ Data Imbalance.} As previously mentioned, we integrate the graphs dataset with the algorithms dataset and use LLMs to generate problems and code for each integration. However, the code produced by LLMs may not always be accurate. After data cleaning, the distribution of data across graph problems remains uneven. Some graph problems have very limited data, posing challenges to the training process.

 \textbf{$\bullet$ Problem Variants.} A single graph problem can have multiple variants. For example, the shortest path problem may refer to the shortest path from a single source to all other nodes, or between two specific nodes. LLM-generated data might only cover one variant, necessitating additional data supplementation.

\textbf{$\bullet$ Algorithm Deficiency.} Some graph problems are not directly supported by NetworkX and require modifications to existing algorithms. For instance, to solve the maximum clique problem, one method is to enumerate the cliques in the graph by using \texttt{nx.find\_cliques} and output one clique whose vertex set size is maximum among all cliques in the graph.

To address these issues, we enhance our dataset by collaborating with expert researchers in the field of graph algorithms. Together, we create additional documents, each encompassing the following elements: \texttt{Problem}, \texttt{Graphtype}, \texttt{Parameters}, and \texttt{Code}. These elements represent the problem description, the type of graph being used, the necessary parameters for the problem, and the code required to solve it. These newly created documents are compiled into a comprehensive algorithm dataset. We employ the same method to seamlessly integrate this additional data with the existing graph dataset. This integration results in a richer collection of training data that covers a broader range of graph computational problems. Following data augmentation, our final \ourdataset dataset contains $N = 49,224$ samples, providing a robust foundation for subsequent model fine-tuning. These efforts significantly enhance the model's capability to understand and solve complex graph-related challenges.

\subsection{Multi-Stage Fine-Tuning} 
Based on the constructed graph code dataset \texttt{GraphWild}, we develop \ourmodel by fine-tuning open-sourced LLMs. Figure \ref{RAG_inference}(a) demonstrates our training pipeline.
The fine-tuning process involves two major techniques: multi-task Supervised Fine-Tuning (SFT) and Reinforcement Learning from Compiler Feedback (RLCF).
SFT enhances the model's tool-using capability for solving diverse graph computational problems. RLCF improves its cognitive ability to distinguish incorrect code from correct code. 

\textbf{$\bullet$ Supervised Fine-Tuning (SFT).} In the supervised fine-tuning stage, we train the model on \ourdataset to teach it how to solve complex problems. The model is trained to generate code $\hat{\mathcal{C}}_i$ based on a given problem $\mathcal{Q}_i$ and its corresponding input graph $\mathcal{G}_i$. The training loss is calculated as the difference between the model's predicted code $\hat{\mathcal{C}}_i$ and the ground-truth code $\mathcal{C}_i$ in the training dataset, formally defined as follows:
\begin{align}
    \mathcal{L}_{\text{SFT}}=-\sum\nolimits_{i=1}^{N}\log \textbf{P}(\hat{\mathcal{C}}_i|\mathcal{Q}_i,\mathcal{G}_i, \theta),
    \label{eq:sft_loss}
\end{align}
where $N$ is the size of our training dataset, and $\theta$ is the parameters of the base model $\mathcal{M}$.
In our experiment, we fine-tune Llama3.1-8b and Qwen2.5-coder with the Low-Rank Adaption (LORA) method ~\cite{hu2021lora}. The fine-tuned model is termed as $\mathcal{M}_{\text{SFT}}$.

\textbf{$\bullet$ Reinforcement Learning from Compiler Feedback (RLCF).} In order to refine the model's code generation process to enhance task-specific performance, we propose Reinforcement Learning from Compiler Feedback (RLCF), similar to Reinforcement Learning from Human Feedback (RLHF) but with preferences determined by the compiler instead of humans. Specifically, its objective is to train our model to distinguish between runnable code that produces the correct solution and incorrect code. To achieve this, we construct a training corpus using the following method: for each problem, the trained model $\mathcal{M}_{\text{SFT}}$ performs inference 100 times, generating codes with corresponding prompt $\{\mathcal{Q, G}\}$. Each code is executed, and the result is compared to the ground truth. Runnable code that provides the correct answer performs the chosen item, while code yielding incorrect results performs the rejected item. The paired sample chosen can be denoted as:
\begin{align}
    \mathcal{C}_{t} \succ \mathcal{C}_{f} | (\mathcal{Q, G}), \quad \{(\mathcal{Q,G}), \mathcal{C}_{t}, \mathcal{C}_{f}\} \sim \mathcal{D}_{\text{RLCF}},
\end{align}
where $\mathcal{C}_{t}$ and $\mathcal{C}_{f}$ denote the preferred and dispreferred codes respectively. RLCF directly optimizes the policy that best satisfies preferences through a simple categorical objective, fitting an implicit reward model, and the corresponding optimal policy can be extracted in closed form. The parametrized policy model $\mathcal{M}_{\theta}$ is optimized with a maximum likelihood objective:
\begin{align}
\begin{split}
    &\mathcal{L}_{\text{RLCF}}(\mathcal{M}_{\theta}; \mathcal{M}_{\text{SFT}}) = -\mathbb{E}_{(\mathcal{Q, G};\mathcal{C}_{t}, \mathcal{C}_{f})\sim \mathcal{D}_{\text{RLCF}}} \\
    &\left[ \log \sigma\left(\beta \log \frac{\mathcal{M}_{\theta}(\mathcal{C}_{t}|\mathcal{Q,G})}{\mathcal{M}_{\text{SFT}}(\mathcal{C}_{t}|\mathcal{Q,G})}-\beta\log\frac{\mathcal{M}_{\theta}(\mathcal{C}_{f}|\mathcal{Q,G})}{\mathcal{M}_{\text{SFT}}(\mathcal{C}_{f}|\mathcal{Q,G})}\right)\right],
\end{split}
\end{align}
where $\mathcal{D}_{\text{RLCF}}$ is our selected dataset with $\{(\mathcal{Q,G})-\mathcal{C}_{t}-\mathcal{C}_{f}\}$ for policy optimization, we collect 3,000 samples for our RLCF training corpus. $\beta$ is a scaling factor that controls the extent to which the implicit reward model is incorrect about the complementary ordering, accounting for the strength of constraint.



\begin{table*}[th]
    \centering
    \resizebox{\textwidth}{!}{
    \begin{tabular}{c|ccccccccc}
    \toprule
       Task & GPT-4o & GPT-4o-mini & Qwen2.5 coder & Llama3.1-8b  & Graphwiz & \texttt{GCoder}-Q & \texttt{GCoder}-L \\
    \midrule
    
    Bipartite  & 99.25 & \underline{99.75} & 92.75 & 24.50 & 72.00  & 95.75 & \textbf{100.00} \\
    Topology Sort  & 97.75 & 94.50 & 83.50 & 4.25 & 34.50  & \underline{99.00} & \textbf{99.50} \\
    Shortest Path  & 93.50 & 73.50 & \underline{97.00} & 6.75 & 13.00  & 77.00 & \textbf{100.00} \\
    Hamilton Path &68.00 & 59.25 & 70.75 & 62.33 & 38.00 & \underline{72.00} & \textbf{98.75} \\
    Maximum Flow  & 43.33 & 48.00 & 22.00 & 7.25 & 48.75  &\textbf{93.67} & \underline{81.00} \\
    Clustering Coefficient  & 62.67 & 50.33 &68.66 & 52.00 & 21.00 & \textbf{78.33} & \underline{78.30} \\
    Common Neighbor  & \underline{98.14} & 64.65 &96.27 & 98.14  & 18.60 & \textbf{99.53} & 95.81 \\
    Connected Component  &52.33 & 33.33 &30.33 & 5.67 & 2.33 & \underline{82.67} & \textbf{83.30} \\
    Connectivity  & \textbf{100.00} & 91.33 &78.50 & 89.67 & 51.00  &\textbf{100.00} & \textbf{100.00} \\
    Euler Path  &75.00 & 66.33 &78.00 & 49.25 & 48.33 & \textbf{100.00} & \textbf{100.00} \\
    Diameter  & 82.00 & 94.00  &98.00 & 43.00 & 11.67  & \textbf{100.00} & \textbf{100.00} \\
    Regular  & \textbf{100.00} & \textbf{100.00}  &99.75 & 68.75 &  47.75 & 99.00 &  96.25 \\
    Distance Regular  & \textbf{99.75} & 79.50  & 85.25 & 69.75 & 50.25  & 99.25 & \textbf{99.75} \\

    Overall &81.08 &73.00 &76.29 &42.63 & 35.17 & \underline{91.27} & \textbf{94.39} \\

    \bottomrule    
    
    \end{tabular}}
    \caption{Performance of \ourmodel and other baselines on our test sets. \texttt{GCoder}-Q and \texttt{GCoder}-L represent our model fine-tuned based on Qwen2.5-coder and Llama3.1-8b, respectively. The \pmb{best} is highlighted in bold, and the \underline{second-best} is underlined.}
    \vspace{-7mm}
    \label{tab:main_results}
\end{table*}

\subsection{RAG-enhanced Inference} \label{RAG inference}

After fine-tuning the model with our \ourdataset dataset, we reach the final stage of our pipeline: conducting inference on \ourmodel to generate high-quality code for the queried graph problem. Unlike previous work that mainly relies on in-context learning to enhance LLMs' performance on graph computational tasks, our approach uses \textbf{Retrieval Augmentation Generation (RAG)} technique based on LangChain~\cite{LangChain}, through which documents can be retrieved back to \ourmodel to improve the transferability to unseen tasks of \ourmodel as well as automate the entire process of adding examples. Figure \ref{RAG_inference}(b) demonstrates our inference pipeline, which consists of the following key steps: \texttt{Domain Classification}, \texttt{Code Library Construction}, \texttt{Indexing}, \texttt{Retrieval} and \texttt{Generation}.

\noindent\textbf{Domain Classification.} Although the correct documents retrieved by RAG techniques can enhance \texttt{GCoder}'s performance on graph tasks not included in \ourdataset (we call them \texttt{Out-of-Domain} tasks), several issues arise with the use of RAG: (1) The retrieval process incurs additional time overhead, particularly when the retrieval library is large, increasing user wait times. (2) \ourmodel already performs well on tasks contained in the training dataset (We call them \texttt{In-Domain} tasks), and retrieved documents might be incorrect, potentially degrading performance on these tasks. To address this, we perform domain classification by maintaining a list of in-domain tasks and checking if the query matches these tasks. We then conduct direct inference for in-domain tasks and RAG-enhanced inference for out-of-domain tasks, respectively.

\noindent\textbf{Code Library Construction.} A comprehensive and accurate information retrieval library can expand the range of tasks \ourmodel can handle while ensuring accuracy. To achieve this, we construct the code library from our training dataset and code designed by experts. These codes are stored in a CSV file, along with the corresponding task name. Each row of the CSV file corresponds to one data.

\noindent\textbf{Indexing.} We calculate the length for each data in the code library and set the longest one as the chunk size. Then the code library is segmented into smaller chunks, where each chunk represents
a piece of data. The chunks are subsequently embedded as vectors using BAAI/bge-base-en-v1.5~\cite{xiao2023c}. The vectors are then stored in the \texttt{Chroma} vector database.

\noindent\textbf{Retrieval.} One challenging issue is retrieving the most relevant code for a given graph task. This difficulty arises because many similar tasks exist in the code library, such as the maximum clique problem and the maximal clique problem. To address this, we combine similarity search with keyword search. Specifically, for a user query, we first apply a similarity search based on the query and data to filter out irrelevant data. Then, we perform a keyword search using the task name of each data entry to accurately match the query and distinguish the most relevant data entries from similar entries. Retrieved documents are appended to the original query.

\noindent\textbf{Generation.} The query is sent to \ourmodel to generate code, which is executed by external tools. The execution result provides the answer to the query.

\section{Experiments}
In this section, we evaluate \ourmodel by addressing the following research questions:

\noindent\textbf{RQ1}: How does \ourmodel perform across different in-domain graph tasks, and how does it compare to other powerful LLMs?

\noindent\textbf{RQ2}: How does \ourmodel handle out-of-domain (OOD) tasks?

\noindent\textbf{RQ3}: How does \ourmodel perform as the graph size increases, and how large graph can \ourmodel handle?

\noindent\textbf{RQ4}: What is the ability of \ourmodel to handle diverse input formats?

\begin{figure*}[t]
    \centering
    \includegraphics[width=\textwidth]{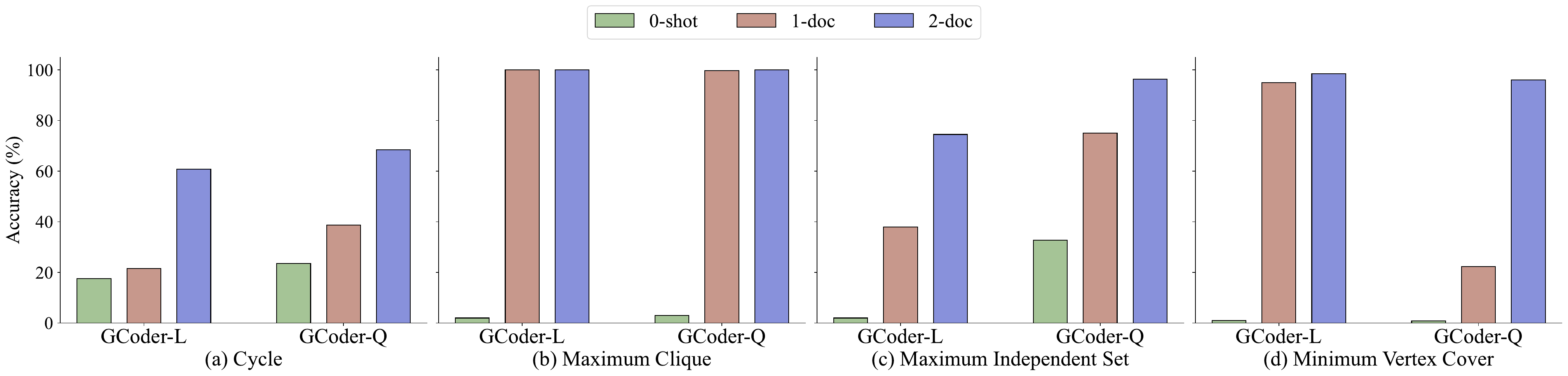}
    \caption{RAG boosts the performance of out-of-domain tasks. Where 0-shot represents direct inference, 1-doc and 2-doc denote the chunk numbers of RAG. We perform RAG inferences with 2 fine-tuned models (i.e., GCoder-L bases on Llama3.1-8b and GCoder-Q bases on Qwen2.5-coder), and results show the effectiveness of RAG.}
    \label{fig:ood_task_rag}
\end{figure*}

\subsection{Experiments Settings}
\noindent\textbf{Baseline LLMs.} The baseline LLMs in our experiments are classified into three categories: closed-source LLMs, open-source LLMs, and graph-task-specific LLMs. For closed-source LLMs, we evaluate GPT-4o and GPT-4o-mini. For open-source LLMs, we evaluate Qwen2.5 Coder and Llama3.1-8b. For graph-task-specific LLMs, we evaluate Graphwiz based on Llama2-13b.

\noindent\textbf{Evaluation Datasets.} To thoroughly evaluate LLMs across a variety of tasks and graph input formats, we utilized data from 4 graph computational datasets: GraphInstruct~\cite{luo2024graphinstruct}, NLGraph~\cite{wang2023can}, Graphwiz~\cite{chen2024graphwiz}, and GraphArena~\cite{tang2024grapharena}. These datasets provide a diverse range of graph-related challenges, ensuring a comprehensive assessment of LLM capabilities. 

In addition to these datasets, we recognized that \ourmodel addresses a wider spectrum of graph problems than those currently available. To bridge this gap, we generated additional graph tasks using a program-aided generator based on the dataset construction method mentioned in Section \ref{Dataset Construction}. This approach ensures that our evaluation encompasses unique and complex graph scenarios, further testing the adaptability and proficiency of the LLMs in handling a broader array of graph-based problems.

\noindent\textbf{Experimental Setup.} In this work, we utilize open-source models Llama 3.1-8b and Qwen2.5-coder as our backbone architectures, enabling us to construct \ourmodel at multiple scales. Our codebase leverages Llamafactory~\cite{zheng2024llamafactory} and the Huggingface Library. For all models, we configure the learning rate to 1e-5, run for 2 epochs, and set the maximum sequence length to 4096, using NVIDIA 4$\times$A800 (80GB) GPUs. The batch size is either 8 or 16. 

\subsection{Main Experiments (RQ1)}
In this experiment, we selected 13 graph computational tasks to evaluate baseline LLMs and \ourmodel. These tasks are categorized into those with parameter input and those without, including polynomial-time and NP-hard tasks, as well as judgment-based and computational problems, the detailed introduction to those tasks is in Appendix \ref{Task Description}. The test data for these tasks is generated using our data generation method, as described in Section \ref{Dataset Construction}. Each evaluation was repeated three times, and we reported the average performance. Additionally, to ensure fairness, Graphwiz is required to generate answers through step-by-step reasoning, while other LLMs are tasked with generating code. Table \ref{tab:main_results} presents the results. We have the following observation: (1) The minimum accuracy achieved by \ourmodel is 78.30\% in the Clustering Coefficient task, while the maximum accuracy is 100\% in tasks such as Bipartite, Connectivity, Euler Path, and Regular. The average accuracy for models based on Qwen2.5-coder is 91.27\%, and for Llama3.1-8b, it is 94.39\%. This indicates a strong capability in handling various tasks, whether they are polynomial-time, NP-hard, judgment-based, or computational. (2) \ourmodel achieves performance comparable to strong open-source models like GPT-4o and GPT-4o-mini. In some tasks, it even surpasses them significantly, demonstrating the effectiveness of our pipeline. (3) \ourmodel performs significantly better than its base models, i.e., Llama3.1-8b and Qwen2.5-coder, highlighting the successful application of our multi-stage fine-tuning techniques, which substantially enhance performance over the base models.

\begin{figure}[t]
    \centering
    \includegraphics[width=\linewidth]{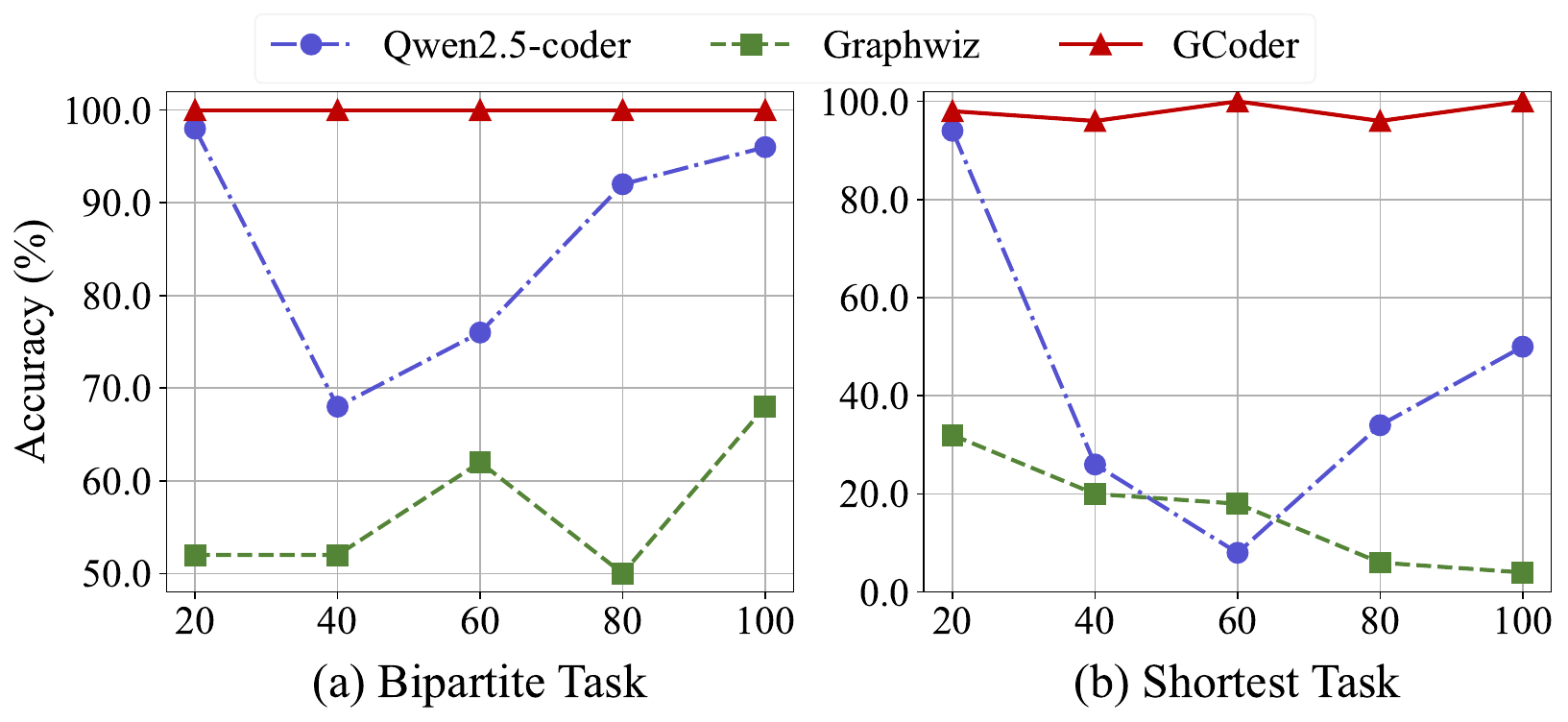}
    \caption{Performance with different graph sizes on Bipartite and Shortest tasks. \ourmodel clearly outperforms Qwen2.5-coder and Graphwiz baselines and achieves stable performance with graph size increasing.}
    \vspace{-4mm}
    \label{fig:graph_size}
\end{figure}

\subsection{\ourmodel on OOD tasks (RQ2)}
In this experiment, we evaluate \ourmodel using the backbone models Llama3.1-8b and Qwen2.5-coder on 4 out-of-domain (OOD) tasks: Cycle Detection, Maximum Clique Problem, Maximum Independent Set Problem, and Minimum Vertex Cover Problem. Additionally, we compare the accuracy of \ourmodel across different settings: 0-shot direct inference, retrieval of 1 document using RAG techniques, and retrieval of 2 documents using RAG techniques. Figure \ref{fig:ood_task_rag} illustrates the results.
We observe that although \ourmodel performs relatively poorly in the 0-shot setting, with the highest accuracy not exceeding 20\%, the use of RAG techniques significantly enhances performance. In particular, accuracy reaches 100\% for both the Maximum Clique Problem and the Minimum Vertex Cover Problem. This demonstrates several key points:
(1) Effective Code Library and Hybrid Search Strategy: The success of both our code library and the hybrid search techniques, as discussed in Section \ref{RAG inference}, is evident. These strategies have proven to be highly effective in enhancing performance.
(2) Improved Transferability: With the enhancement of only 1 or 2 documents, \ourmodel achieves satisfactory performance, indicating strong transferability and adaptability to different tasks.
These results highlight the robustness and versatility of \ourmodel, especially when supplemented with relevant information through RAG techniques.

\begin{figure}[t]
    \centering
    \includegraphics[width=\linewidth]{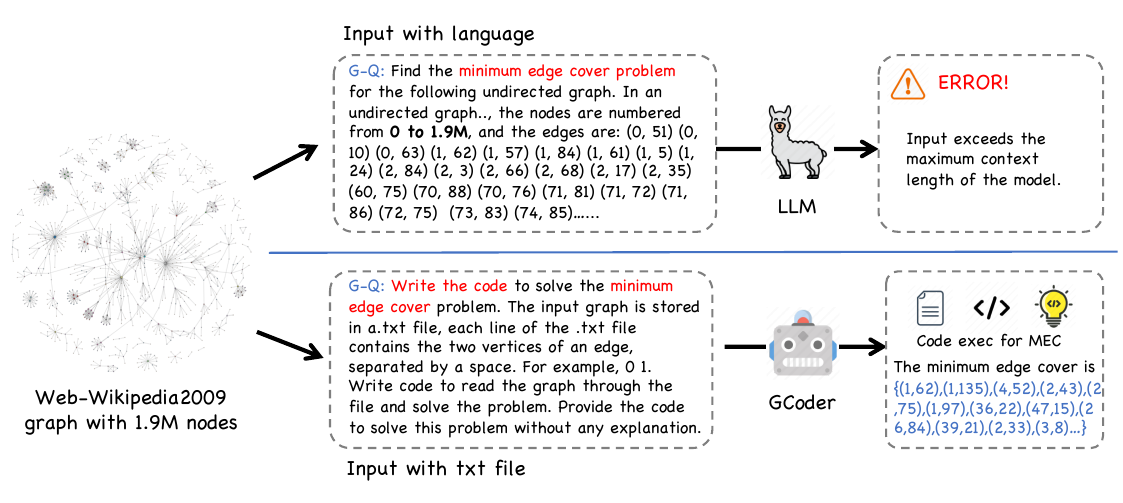}
    \caption{Input with language for reasoning cannot handle large graph, while \ourmodel is free from graph size by writing code and then executing with reading the graph from a file.}
    \vspace{-4mm}
    \label{fig:grah_file}
\end{figure}

\subsection{\ourmodel on Large Graphs (RQ3)}
To evaluate the performance of \ourmodel on large graphs, we conduct two experiments to asses performance variants when graph size changes and a case study with respect to a million nodes graph.



\begin{figure*}[t]
    \centering
    \includegraphics[width=\linewidth]{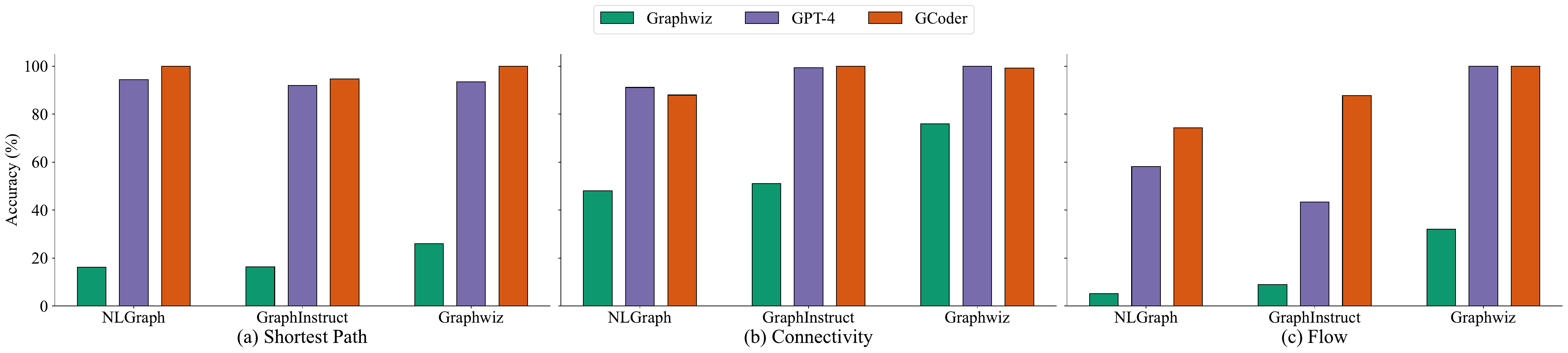}
    \caption{Performance of Graphwiz, GPT-4o, and GCoder on 3 open-sourced datasets: NLGraph, GraphInstruct, and the dataset of Graphwiz. GCoder achieves the highest accuracy, exceeding 60\% across all datasets, outperforming the other LLMs.}
    \vspace{-2mm}
    \label{fig:diverse_graph}
\end{figure*}

\noindent\textbf{Performance Variability of Different LLMs with Changing Graph Size.} In this experiment, we investigated the impact of graph size on performance, focusing on the robustness of \ourmodel compared to others. We tested \ourmodel on two tasks: Bipartite and Shortest Path. Figure \ref{fig:graph_size} illustrates the results. It is evident that \ourmodel consistently achieved nearly 100\% accuracy across all graph sizes, demonstrating its superior stability and effectiveness. In contrast, other models showed significant performance fluctuations and declines as graph size increased.

Graphwiz's difficulties are primarily due to its reliance on reasoning-based approaches, which become problematic with larger graphs. Two main issues arise: (1) As graph size grows, Graphwiz experiences hallucinations when reading graphs. It occasionally extracts non-existent nodes and edges, leading to errors. This misinterpretation significantly contributes to its performance decline. (2) Additionally, larger graphs require longer and more complex natural language reasoning. The longer the natural language reasoning, the more likely a language model is to make errors, increasing cognitive load and leading to inaccuracies.

Conversely, \ourmodel excels in these scenarios due to its code-based problem-solving approach. By generating code, \ourmodel bypasses the need for a complete structural understanding. It simply memorizes the graph and inputs parameters into the code, reducing cognitive demands. Furthermore, the shorter inference path associated with code generation minimizes error potential, providing a clear advantage over reasoning-heavy methods.

Moreover, while both \ourmodel and Qwen2.5-coder employ code-based solutions, \ourmodel is specifically optimized for graph computational problems. This specialization allows it to outperform the Qwen2.5-coder, as evidenced by its consistent near 100\% accuracy across various graph sizes. Our pipeline equips \ourmodel to handle graph-related tasks more effectively, underscoring its superior adaptability and precision.

\noindent\textbf{Case study: \ourmodel's performance on million nodes graph.} In this experiment, we present a case study to test the capability of \ourmodel in handling a million-node graph. We selected a well-known web graph, \texttt{Web-Wikipedia2009}, an undirected graph from Networkrepository~\cite{social}, with 1.9 million nodes and 4.5 million edges. The corresponding problem is the minimum edge cover, which aims to find a set of edges with minimum cardinality such that every vertex is incident to at least one edge of the set.
Due to the numerous edges in the graph, we bypass the input token limitation of \ourmodel by sending only the problem text, prompting it to generate code to extract the graph information from a '.txt' file and solve the problem. We also run a classic solver to obtain the ground truth. Figure \ref{fig:grah_file} demonstrates the difference between sending the whole natural language description of the graph into LLMs and prompting \ourmodel to generate the response based on a file. While the input length exceeds other LLMs' token limitations, \ourmodel can generate code based on the file.
We then execute the code generated by \ourmodel, compare the result with the ground truth, and verify its correctness. From this case study, we observe that \ourmodel can handle larger graphs than previous LLMs designed to solve graph computational problems, which could only handle graphs with up to 100 nodes.


\begin{figure}[t]
  \centering
  \begin{minipage}{0.23\textwidth}
    \includegraphics[width=\textwidth]{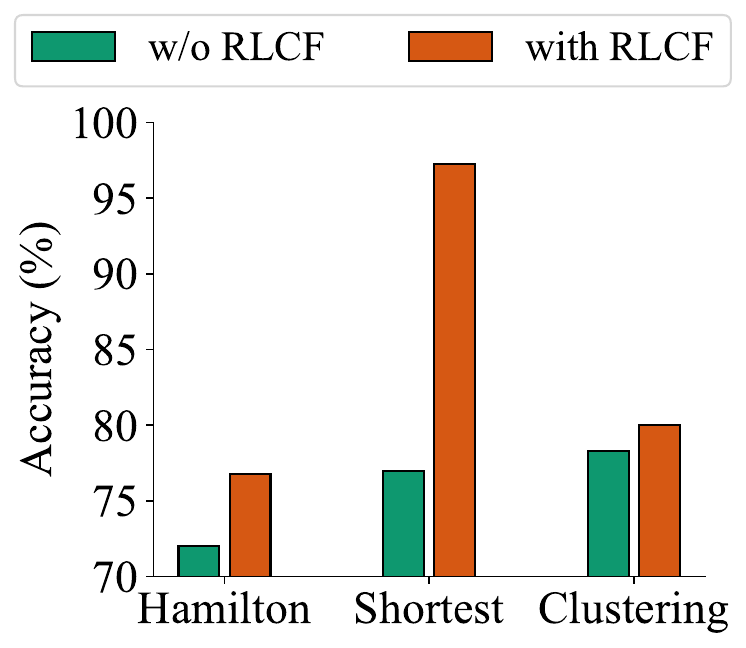}
    \vspace{-4mm}
    \caption{RLCF boots the performance of \texttt{GCoder}.}
    \label{fig:with_rlcf}
  \end{minipage}
  \hfill 
  \begin{minipage}{0.23\textwidth}
    \includegraphics[width=\textwidth]{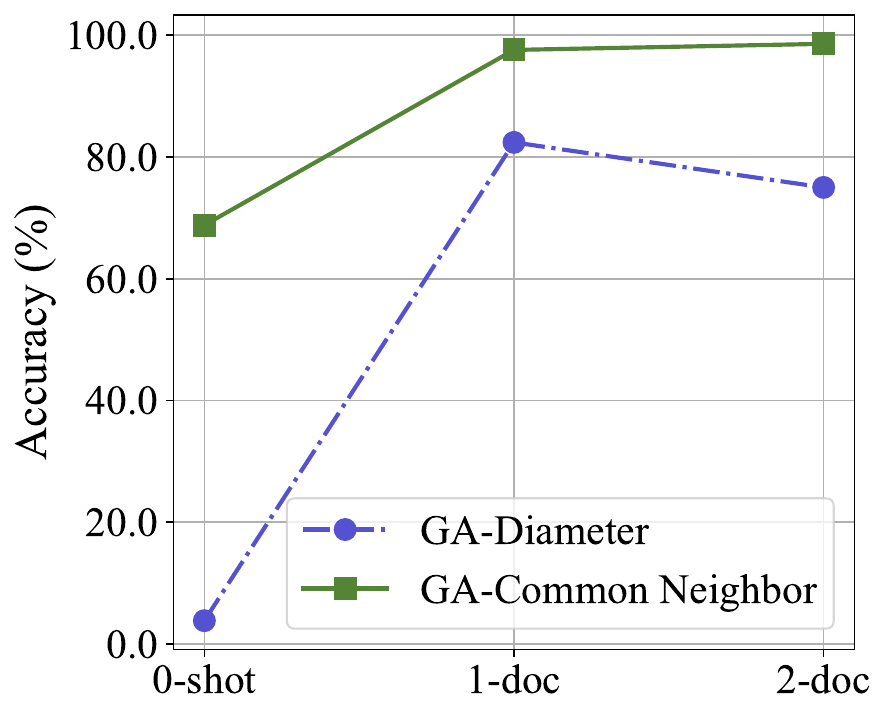}
    \caption{Performance of \ourmodel on Real-Life Graphs.}
    \label{ex:OOD-graph}
  \end{minipage}
  \vspace{-4mm}
\end{figure}

\subsection{\ourmodel on Diverse Graph Input (RQ4)} In this section, we test the capability of \ourmodel on graph tasks of diverse input format, which can be categorized into graphs of traditional data structure and graphs of real-life scenarios. 

\noindent\textbf{Experimentation on Graphs of Traditional Data Structures.} In this experiment, we evaluate three LLMs: \texttt{GCoder}, Graphwiz, and GPT-4o, on three datasets designed for graph computational problems: GraphInstruct, Graphwiz's dataset, and NLGraph. The graphs in these datasets are described using traditional data structures such as edge lists and adjacency lists. We focus on three tasks that are common across these datasets: shortest path, maximum flow, and connectivity. Figure \ref{fig:diverse_graph} illustrates the results.
We can observe that:
(1) \ourmodel performs well across all datasets. This is largely because its training data includes abundant examples of graphs described using traditional data structures.
(2) Graphwiz performs best on its own dataset but does not generalize as well to others. This is likely due to its limited exposure to the input formats found in other datasets.
(3) GPT-4o shows good performance across various datasets. However, it lags behind \ourmodel because it has not been fine-tuned specifically for graph-related tasks.

\noindent\textbf{Experimentation on Graphs of Real-Life Scenarios: } In this experiment, we evaluate \ourmodel on two tasks of GraphArena: Common Neighbor and Diameter, with graph descriptions related to the DBLP academic network and DBPedia knowledge graph. The graph description format is not contained in \texttt{GraphWild}. We test the performance of \ourmodel in the settings of 0-shot direct inference, retrieval of 1-doc and 2-doc using RAG techniques. Figure \ref{ex:OOD-graph} demonstrates the results, from which we can observe that: (1) Thanks to the step-by-step graph reading examples in the training dataset, \ourmodel performs well in the Diameter task, with accuracy reaching almost 70\%. (2) With the enhancement by the RAG technique, \ourmodel achieves a significant improvement, breaking through 70\% accuracy in the Common Neighbor task and reaching almost 100\% accuracy in the Diameter task. (3) We observe a slight decrease in performance for the Common Neighbor task with the 2-doc setting, possibly due to the incompatibility of the retrieved documents.

\begin{figure}[t]
    \centering
    \includegraphics[width=0.98\linewidth]{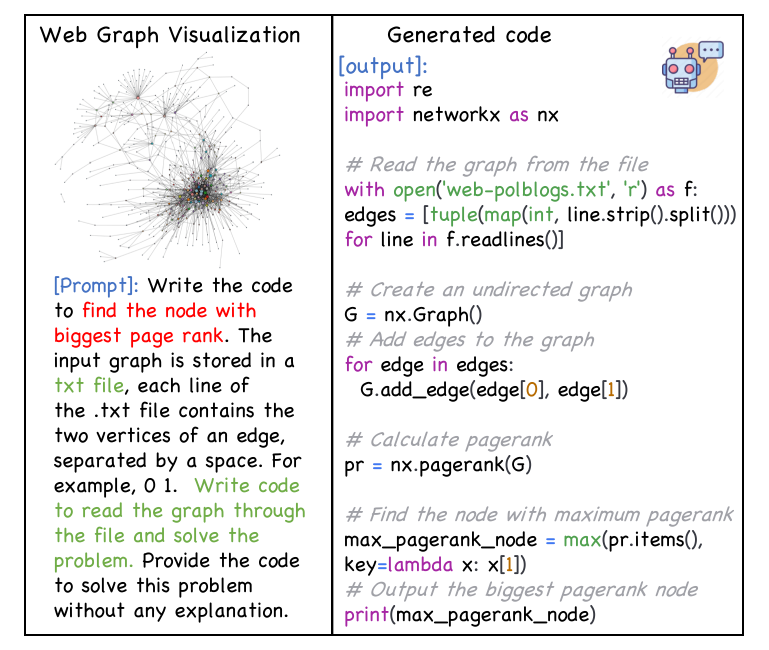}
    \caption{A showcase of hiring \texttt{GCoder} for blog retrieval according to pagerank scores.}
    \label{fig:web_graph}
\end{figure}

\subsection{Ablation Study}
To evaluate the effectiveness of RLCF in enhancing the performance of \ourmodel on tasks with relatively low accuracy, we compared \ourmodel with and without RLCF across three tasks: Hamilton Path, Shortest Path, and Clustering Coefficient. Figure \ref{fig:with_rlcf} illustrates the results. Incorporating RLCF led to accuracy improvements of 5\%, 26\%, and 2.17\% for these tasks, respectively, showing that \ourmodel can better identify and optimize its output to achieve superior performance through the reward mechanism of RLCF.
\subsection{\ourmodel Helps Blog Retrieval with Pagerank}
PageRank is a ranking algorithm used by search engines to assess the importance of web pages by analyzing the number and quality of links, effectively prioritizing search results. In this section, we apply the PageRank algorithm to \texttt{Web-Polblogs}, a web graph consisting of 643 nodes and 2,280 edges. Each node represents a blog webpage, and edges indicate hyperlinks between these blogs, reflecting their interconnectedness.
Figure \ref{fig:web_graph} illustrates the prompt provided to \ourmodel and the corresponding output, showcasing its capability to handle complex graph data. By leveraging \texttt{GCoder}, we determine the blog with the highest PageRank score, thereby identifying the most influential blog within the network.

\section{Related Works}
\subsection{LLMs on Graphs} 
Large Language Models (LLMs) offer robust capabilities to enhance graph analytics and learning, especially for predictive tasks in text-attributed networks ~\cite{fatemi2023talk, wang2023can, chen2024exploring, chen2024labelfree, tang-etal-2023-fused,huang2023prompt,duan2023frustratingly,xia2024opengraph,li2023survey}. Compared to machine learning tasks on graphs, such as node classification ~\cite{chen2024labelfree}, graph computational problems present greater challenges for LLMs. These challenges arise from the need for a deeper understanding of structural information and long-term, multi-step reasoning.
In this domain, GraphWiz ~\cite{chen2024graphwiz} introduced an instruction-tuning dataset to guide LLMs' reasoning path on graph computation problems. GraphInstruct ~\cite{luo2024graphinstruct} evaluated LLMs' graph understanding, facilitated the development of GraphLM and GraphLM+, and demonstrated their effectiveness in graph reasoning tasks. GraphLLM ~\cite{chai2023graphllm} and GraphGPT ~\cite{tang2024graphgpt} further explored methods to enhance LLM performance in graph-related tasks by integrating encoded information from graph neural networks (GNNs).
Additionally, several benchmarks ~\cite{guo2023gpt4graph, zhang2023llm4dyg, li2024visiongraph, baldassarre2021graphqa, wang2023can, liu2023evaluating, yamada2023evaluating,li2024can,tang2024grapharena,dai2024large,wu2024grapheval2000,yuan2024gracore} have been developed to assess LLMs' ability to solve graph problems.

\subsection{Code-Augmented LLMs}
Augmenting LLMs with code can largely alleviate LLMs’ limitations and improve their reasoning capabilities~\cite{gou2023tora,das2024mathsensei,kalyanpur2024llm}. Recent works boosted LLMs’ generalization with programming as thought processes~\cite{chen2022program,yao2023react,mialon2023augmented,parisi2022talm}, or integrating tool retrievers~\cite{borgeaud2022improving,shao2023enhancing}.
MAmmoTH~\cite{yue2023mammoth} proposed MathInstruct dataset for math reasoning, excels in math problem-solving by combining chain-of-thought and program-of-thought rationales helping LLMs use external tools. CRITIC~\cite{gou2023critic} enabled LLMs to validate and improve their outputs by interacting with tools, mimicking human validation processes. TORA~\cite{gou2023tora} is a tool-integrated reasoning framework that significantly improves mathematical problem-solving by combining natural language reasoning with external computational tools, achieving substantial performance gains on various datasets. Xie et al.~\cite{xie2023decomposition} introduced a step-wise self-evaluation mechanism with stochastic beam search to enhance LLM reasoning accuracy and robustness, achieving significant improvements in benchmark tests. Inspired by these works, we utilize code to enhance LLM's performance on graph computational problems.

\section{Conclusion}
\label{sec:conclusion}
In this work, we introduce \texttt{GCoder}, a code-based LLM designed to overcome the limitations of reasoning-based approaches in graph computational problems. We fine-tuned it using multi-stage training techniques on our comprehensive dataset, \texttt{GraphWild}, which includes hundreds of algorithms and various graph input formats. We further enhance \texttt{GCoder}'s transferability with RAG techniques. Extensive experiments show that \texttt{GCoder} achieves high accuracy across a variety of tasks and scales effectively to large graphs and diverse graph input formats.




\bibliographystyle{ACM-Reference-Format}
\bibliography{sample-base}

\input{appendix}

\end{document}

%% file: intro1.tex
\section{Introduction}

Graph computational problem is a branch of mathematical reasoning that has been extensively studied in areas like social networks \cite{ceccarello2024fast,oettershagen2024finding}, transportation systems\cite{ahmadian2024extracting}, and web mining \cite{pang2024similarity, chen2024link, miyauchi2024local} (e.g., PageRank plays an important role in web retrieval).
Conventional solutions require specialized algorithms and expertise, which limits their broad applicability. Recently, the rise of Large Language Models (LLMs) has introduced a new trend in graph computational solving \cite{guo2023gpt4graph,zhang2023llm4dyg, li2024visiongraph,chen2024graphwiz,luo2024graphinstruct,zhang2024can,wang2024instructgraph}. LLMs offer a more accessible approach by allowing users to input queries in natural language, making complex problem-solving more intuitive and user-friendly across various domains. 

\begin{figure}[t]
    \centering
    \includegraphics[width=0.98\linewidth]{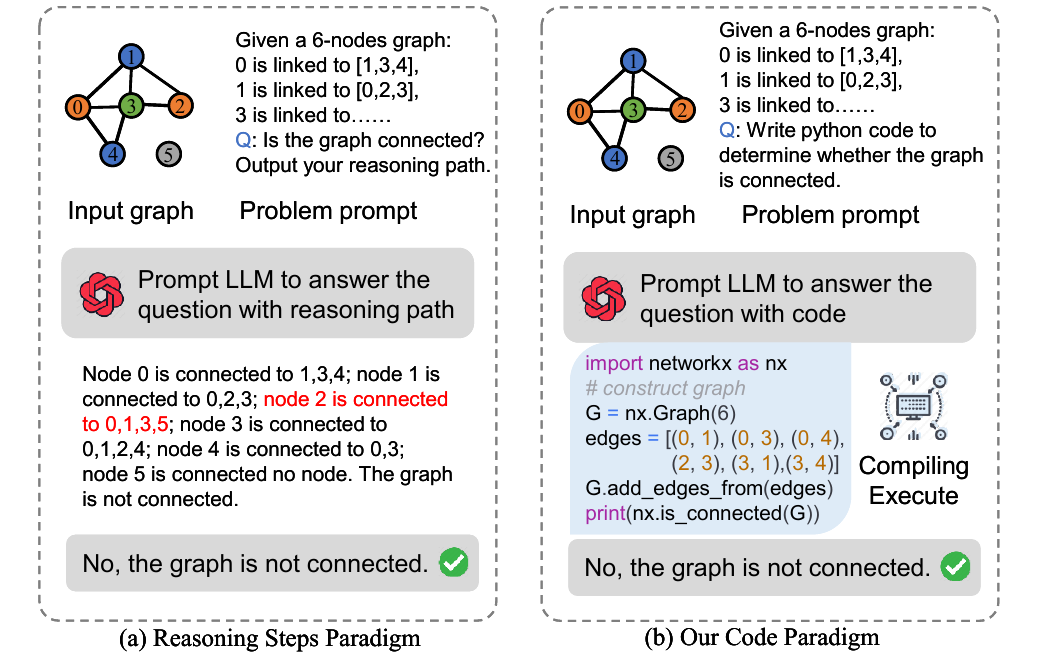}
    \caption{(a) While the reasoning step paradigm outputs correct results, intermediate reasoning can be wrong (i.e., \textcolor{red}{red} reasoning step, node 2 is not connected to 0 and 5). (b) Our code paradigm processes graph problems with programming. More examples can be found in Appendix \ref{sec:QA_case}.}
    \vspace{-4mm}
    \label{fig:intro_demo}
\end{figure}

The ability of large language models (LLMs) to tackle graph computational problems is largely due to their strong reasoning capabilities. Recent research exploits this by using techniques such as chain-of-thought (CoT), where LLMs break down complex problems into a series of simpler, logical steps to solve graph-related challenges \cite{chen2024graphwiz,luo2024graphinstruct}.
However, there are some challenges associated with this paradigm:
(1) \textbf{Unverifiable reasoning steps}, the reasoning steps provided by LLMs cannot be directly verified and their response quality can only be judged by the final answer, while errors may occur in the intermediate steps \cite{chen2024graphwiz}. For example, Figure \ref{fig:intro_demo}(a) demonstrates that although the LLM's final output is correct, there are errors in the intermediate steps.
(2) \textbf{Limited in long-term reasoning}, LLMs face challenges with long-term reasoning in graph computation tasks, which involve extensive enumeration and backtracking. These tasks require an understanding of the topological structure's long-range logic \cite{besta2024graph}. The language reasoning approach, which explores all reasoning paths, struggles with problems needing long-range reasoning, especially in large-scale graphs. (3) \textbf{Unsatisfactory generalization}, these reasoning steps methods customize a reasoning path for each pair of query graphs and task, making it highly sensitive to the type and organization of input graphs. Due to the misalignment between structured graph data and textual space, even minor perturbations in graph structure or changes in the problem query paradigm require LLMs to regenerate their reasoning paths. This process is inefficient and results in unstable performance. 

To overcome these limitations, we replace the reasoning steps paradigm with the code paradigm and introduce \texttt{GCoder}, a code-based LLM for generalized graph problem-solving. Unlike previous approaches that fine-tuned models using natural language reasoning, we focus on code-based fine-tuning. Our pipeline begins with constructing a training dataset, \ourdataset, which is built from a graph dataset featuring diverse formats and algorithm datasets sourced from reputable graph algorithm sites like NetworkX. We then employ a multi-stage training process, incorporating Supervised Fine-Tuning (SFT)~\cite{hu2021lora} and Reinforcement Learning from Compiler Feedback (RLCF), similar to Reinforcement Learning from Human Feedback (RLHF)~\cite{bai2022training} but with preferences determined by the compiler instead of humans, to enhance the model's code-writing skills and reduce common errors. Finally, we conduct inference using two methods. For tasks within our training dataset, we directly prompt the model. For unseen tasks, we utilize a hybrid retrieval technique, combining coarse-grained similarity search with fine-grained keyword search, to extract relevant documents from our designed code library and append it to the original prompt for inference. Experiments show that this retrieval-augmented method significantly boosts the performance of \ourmodel on unseen tasks.

Extensive experiments show that our pipeline has the following advantages:
(1) \textbf{Deterministic and verifiable code}, code can be easily verified by executing it and comparing the output with the ground truth, unlike reasoning steps which are not directly verifiable. This reduces the reliance on manually crafted labels, potentially enhancing performance.
(2) \textbf{Extensive and diverse tasks}, compared to the reasoning steps paradigm, code is often more refined, significantly reducing the LLMs' reasoning difficulty, thus suitable for some complex tasks and large graphs that previous LLMs cannot handle. 
(3) \textbf{Excellent transferability}, for a given task, a single code template can be applied to multiple graphs, increasing its applicability to graph-related problems. Additionally, using RAG techniques can further enhance the performance of \ourmodel on unseen tasks. 


The contribution of our work can be summarized as follows:
\begin{itemize}
    \item We collect code-based instruction-tuning datasets specifically designed for solving graph computational problems. This training dataset encompasses hundreds of graph algorithm codes as well as dozens of graph input formats, significantly enhancing models' capability for solving generalized graph computational problems.

    \item We introduce \texttt{GCoder}, a code-based large language model designed to solve graph computational problems. Experimental results show that \ourmodel outperforms GPT-4o with an average accuracy improvement of 16.42\%, when evaluated on various graph tasks.

    \item \ourmodel is also capable of efficiently handling graphs with millions of nodes and supporting a wide range of input formats with high accuracy, overcoming the limitations of previous LLMs based on reasoning steps paradigm.

\end{itemize}

%% file: appendix.tex
\appendix

\clearpage

\section{Tasks Description}\label{Task Description}
In this section, we give a detailed definition of each task mentioned in our experiments.

\noindent\textbf{Bipartite Graph Check.} This task is to determine if a directed graph $\mathcal{G}=\{\mathcal{V},\mathcal{E}\}$ is bipartite. A graph is considered bipartite if its nodes can be split into two distinct sets $\mathbf{U}$ and $\mathbf{V}$ such that no two nodes within the same set are adjacent

\noindent\textbf{Topology Sort.} In an undirected graph $\mathcal{G}=\{\mathcal{V},\mathcal{E}\}$, the task is to detect the existence of a cycle. A cycle can be defined as a sequence of vertices $v_1, v_2, \dots, v_k$ with $k\geq 3$, that forms a closed loop, meaning $v_1=v_k$. Additionally, for all $1\leq i < k$, each vertex $v_i$ must be distinct from the others, and there must be an edge connecting $v_i$ to $v_{i+1}$. 

\noindent\textbf{Shortest Path.} The task requires identifying the shortest path between two nodes in an undirected, weighted graph $\mathcal{G}=\{\mathcal{V}, \mathcal{E}, w\}$, where $w: \mathcal{E} \to \mathbb{R}^+$ assigns a positive weight to each edge. The goal is to find a path connecting the two nodes such that the total sum of the edge weights along this path is minimized.

\noindent\textbf{Hamilton Path.}
This task is to determine whether there is a Hamilton path in a directed graph $\mathcal{G}=\{\mathcal{V},\mathcal{E}\}$. A Hamilton path is defined as a path that traverses each node in the graph exactly once.

\noindent\textbf{Maximum Flow.} Consider a directed, weighted graph $\mathcal{G}=\{\mathcal{V},\mathcal{E}, c\}$, 
where $c: \mathcal{E} \to \mathbb{R}^+$ is a function assigning a positive capacity to each edge, representing the maximum flow that the edge can support. Given a source node $v_s$ and a sink node $v_t$ in $\mathcal{G}$, the task is to devise a plan to maximize the flow from the source $s$ to the sink $t$. 

\noindent\textbf{Clustering Coefficient.} Consider an undirected graph $\mathcal{G}=\{\mathcal{V},\mathcal{E}\}$. The task is to compute the clustering coefficient $v$, which measures the degree to which nodes cluster together. For a node $v$, the local clustering coefficient is:

\[ C(v) = \frac{2 \cdot \bigtriangleup(v)}{\deg(v) \cdot (\deg(v) - 1)} \]
, where $\bigtriangleup(v)$ is the number of triangles through $v$, deg($v$) is the degree of $v$.

\noindent\textbf{Common Neighbors.} Given an undirected graph $\mathcal{G}=\{\mathcal{V},\mathcal{E}\}$ and two nodes $u, v \in \mathcal{V}$, the task is to compute the set of common neighbors. The common neighbors are defined as nodes that are directly connected to both $u$ and $v$.

\noindent\textbf{Strongly Connected Component.} In a directed graph $\mathcal{G}=\{\mathcal{V},\mathcal{E}\}$, a strongly connected component (SCC) is a maximal subgraph where every pair of nodes $u$ and $v$ are reachable from each other. The task is to identify all SCCs in $\mathcal{G}$.

\noindent\textbf{Connectivity.} Given an undirected graph $\mathcal{G}=\{\mathcal{V},\mathcal{E}\}$ and two nodes $u, v \in \mathcal{V}$, the task is to determine if there exists a path connecting $u$ and $v$. 

\noindent\textbf{Euler Path.} In an undirected graph $\mathcal{G}=\{\mathcal{V},\mathcal{E}\}$, an Euler path is a trail that visits every edge exactly once. The task is to determine if such a path exists. An Euler path exists if and only if the graph is connected and exactly zero or two vertices have odd degree.

\noindent\textbf{Diameter.} In an undirected graph $\mathcal{G}=\{\mathcal{V},\mathcal{E}\}$, the diameter is the longest shortest path between any two nodes. The task is to compute this maximum distance to understand the graph's extent.

\noindent\textbf{Regular Graph Check.} In an undirected graph $\mathcal{G}=\{\mathcal{V},\mathcal{E}\}$, a regular graph is one where each vertex has the same degree. The task is to verify if $\mathcal{G}$ is regular by checking if all vertices have identical degrees.

\noindent\textbf{Distance-Regular Graph Check.} In an undirected graph $\mathcal{G}=\{\mathcal{V},\mathcal{E}\}$, a distance-regular graph has the same number of vertices at each distance for any pair of vertices. The task is to determine if $\mathcal{G}$ is distance-regular by verifying this property for all vertex pairs.

\section{Q-A Cases of \ourmodel}
\label{sec:QA_case}
We demonstrate some exmaples of Q-A cases of \ourmodel, showing its capability of handling diverse tasks and various graph input formats.
\begin{figure}[h]
    \centering
    \includegraphics[width=\linewidth]{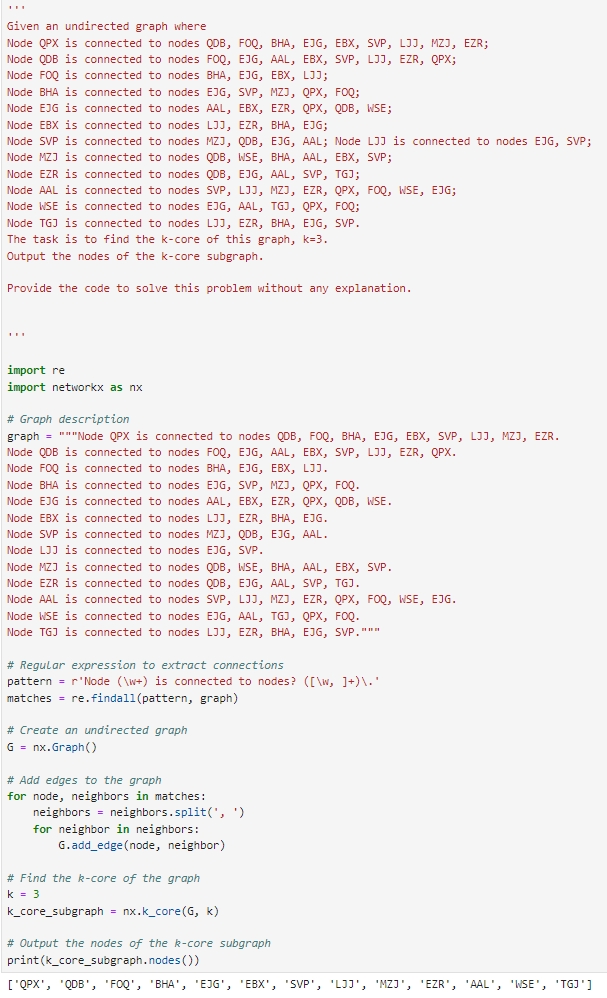}
    \caption{A showcase of k-core.}
    \label{fig:kcore}
\end{figure}


\begin{figure}[ht]
  \centering
  \begin{minipage}{0.45\textwidth}
    \includegraphics[width=\textwidth]{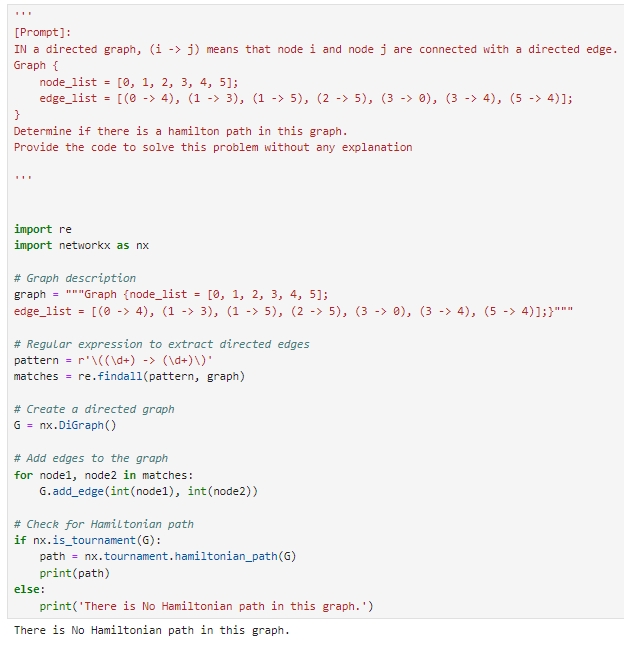}
    \caption{A showcase of Hamilton Path.}
    \label{fig:app_hamilton}
  \end{minipage}
  \hfill 
  \begin{minipage}{0.45\textwidth}
    \includegraphics[width=\textwidth]{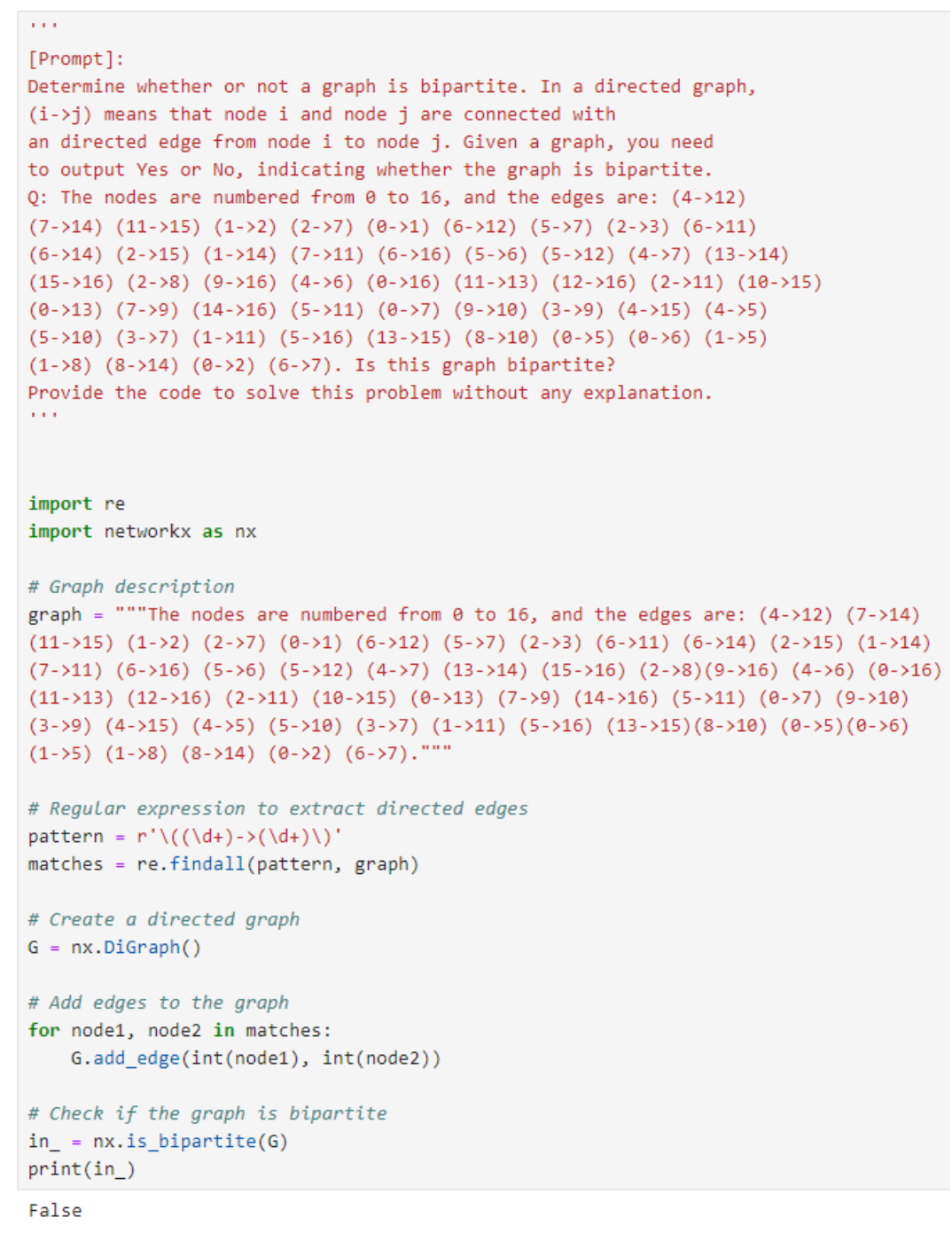}
    \caption{A showcase of Bipartite Task.}
    \label{fig:app_bipartite}
  \end{minipage}
\end{figure}

\begin{figure}[ht]
  \centering
  \begin{minipage}{0.45\textwidth}
    \includegraphics[width=\textwidth]{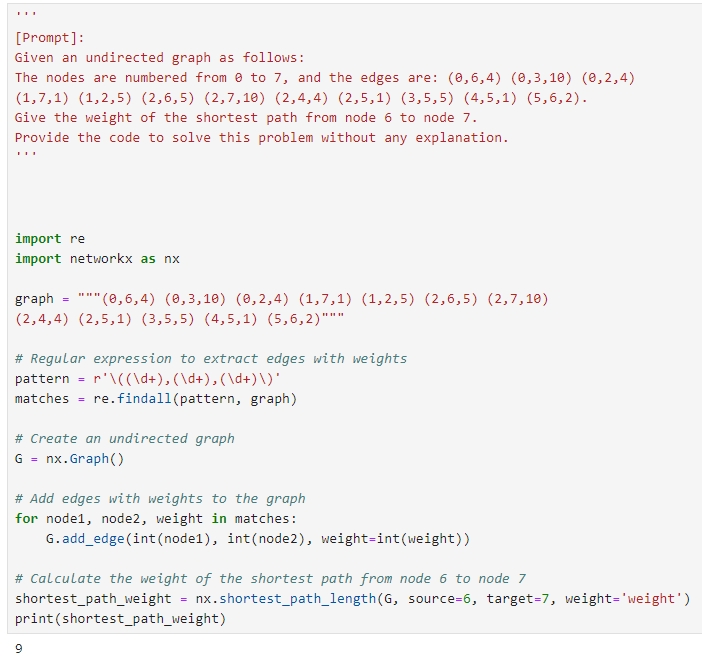}
    \caption{A showcase of Shortest Path Length.}
    \label{fig:app_shortest}
  \end{minipage}
  \hfill 
  \begin{minipage}{0.46\textwidth}
    \includegraphics[width=\textwidth]{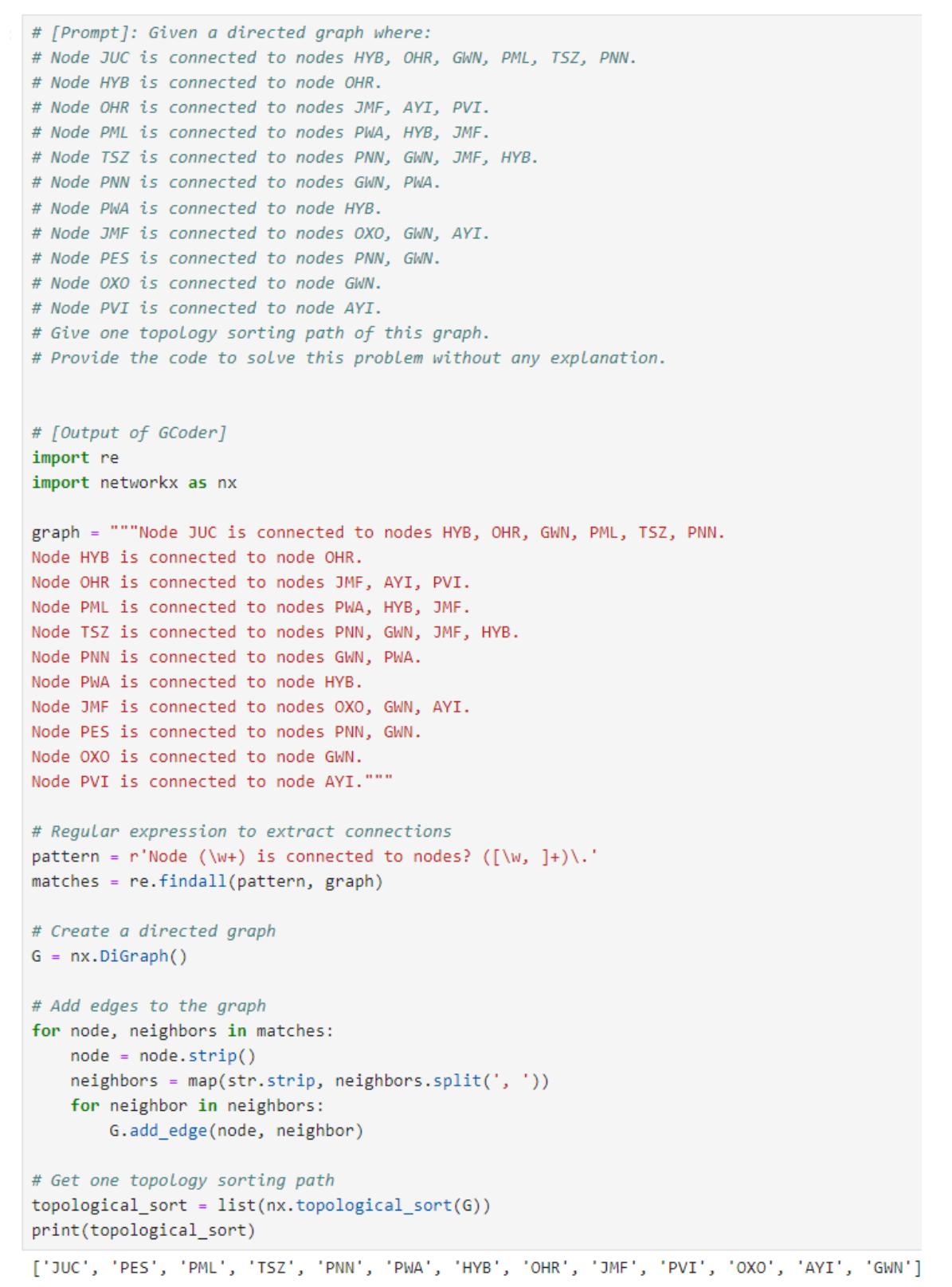}
    \caption{A showcase of Topology Task.}
    \label{fig:app_topology}
  \end{minipage}
\end{figure}

\clearpage

\begin{figure}[h]
    \centering
    \includegraphics[width=\linewidth]{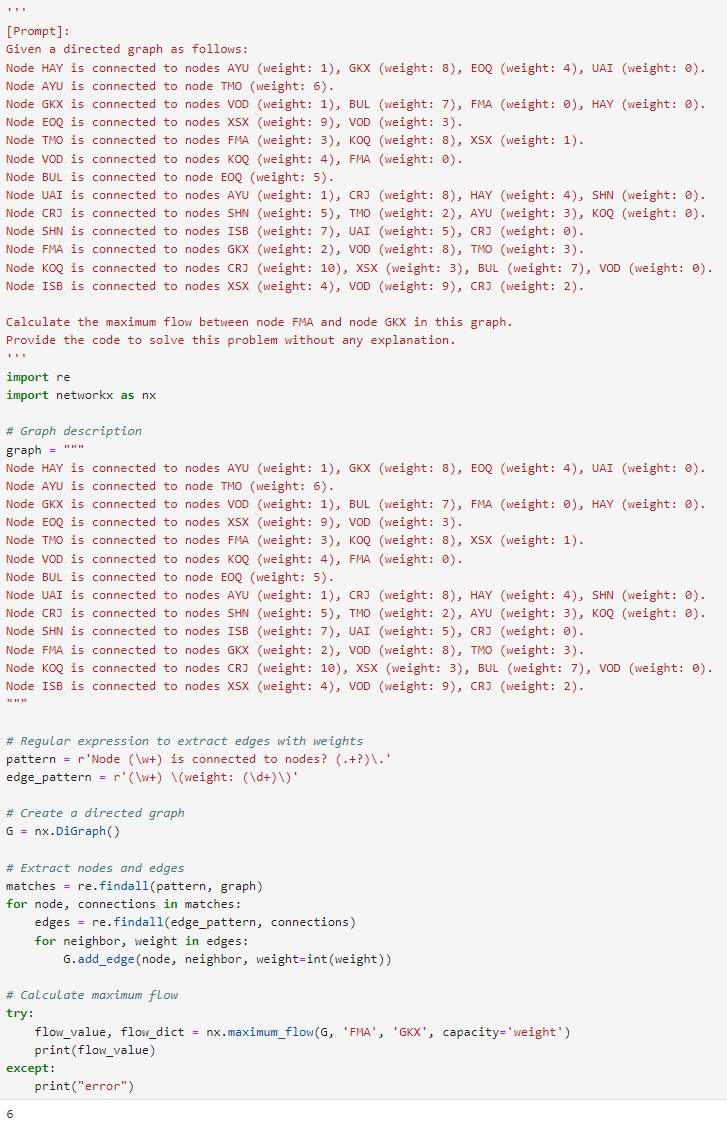}
    \caption{A showcase of Maximum Flow.}
    \label{fig:app_flow}
\end{figure}

\begin{figure}[h]
  \centering
  \begin{minipage}{0.42\textwidth}
    \includegraphics[width=\textwidth]
    {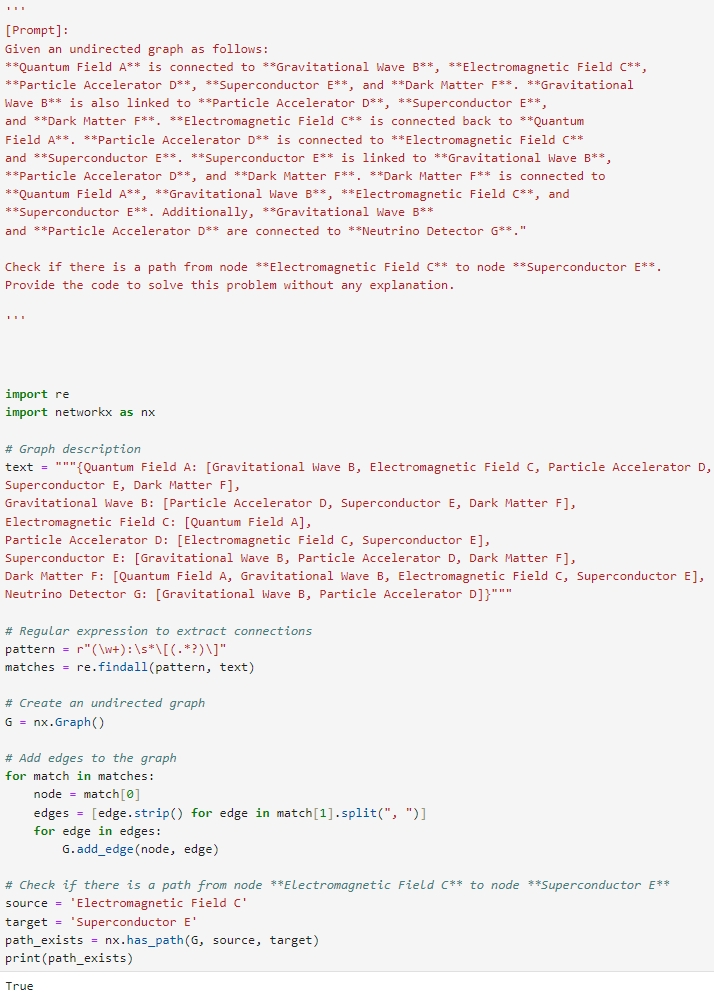}
    \caption{A showcase of Connectivity.}
    \label{fig:app_connectivity}
  \end{minipage}
  \hfill 
  \begin{minipage}{0.42\textwidth}
    \includegraphics[width=\textwidth]{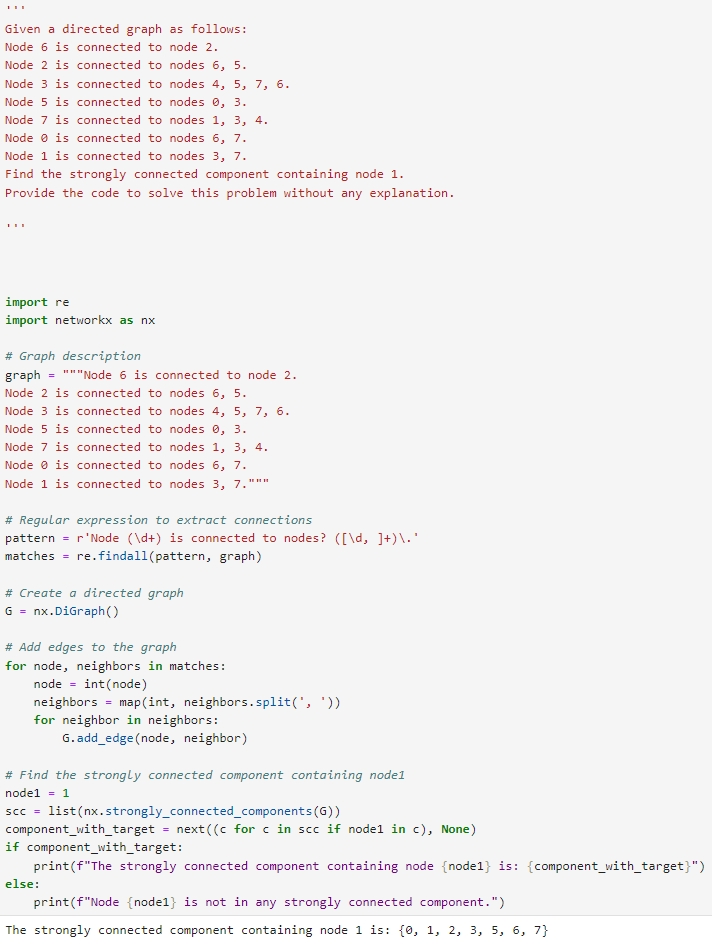}
    \caption{A showcase of Connected Component.}
    \label{fig:app_connected_component}
  \end{minipage}
\end{figure}


\begin{figure}[h]
    \centering
    \includegraphics[width=\linewidth]{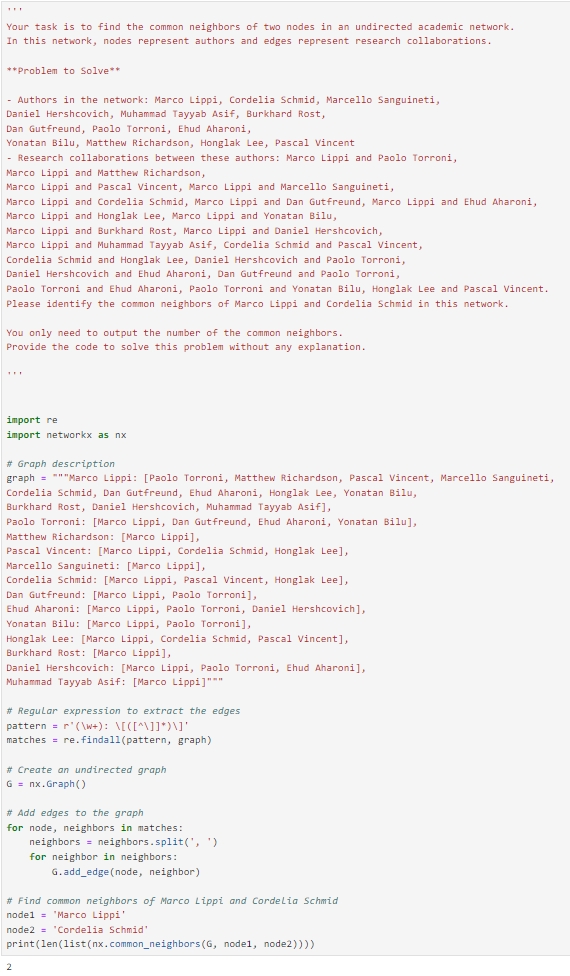}
    \caption{A showcase of Common Neighbor.}
    \label{fig:app_common}
\end{figure}

